\documentclass[conference]{IEEEtran}
\IEEEoverridecommandlockouts

\usepackage{cite}
\usepackage{amsmath,amssymb,amsfonts}
\usepackage{algorithmic}
\usepackage{graphicx}
\usepackage{textcomp}
\usepackage{xcolor}
\def\BibTeX{{\rm B\kern-.05em{\sc i\kern-.025em b}\kern-.08em
    T\kern-.1667em\lower.7ex\hbox{E}\kern-.125emX}}
\begin{document}

\title{Adaptive USVs Swarm Optimization for Target Tracking in Dynamic Environments\\

}

\author{\IEEEauthorblockN{Oren Gal}
\IEEEauthorblockA{\textit{Hatter Department of Marine Technologies} \\
\textit{Charney School of Marine Sciences, University of Haifa}\\
Haifa, Israel \\
orengal@univ.haifa.ac.il}

}

\maketitle

\begin{abstract}
This research investigates the performance and efficiency of Unmanned Surface Vehicles (USVs) in multi-target tracking scenarios using the Adaptive Particle Swarm Optimization with k-Nearest Neighbors (APSO-kNN) algorithm. The study explores various search patterns—Random Walk, Spiral, Lawnmower, and Cluster Search—to assess their effectiveness in dynamic environments. Through extensive simulations, we evaluate the impact of different search strategies, varying the number of targets and USVs' sensing capabilities, and integrating a Pursuit-Evasion model to test adaptability. Our findings demonstrate that systematic search patterns like Spiral and Lawnmower provide superior coverage and tracking accuracy, making them ideal for thorough area exploration. In contrast, the Random Walk pattern, while highly adaptable, shows lower accuracy due to its non-deterministic nature, and Cluster Search maintains group cohesion but is heavily dependent on target distribution. The mixed strategy, combining multiple patterns, offers robust performance across varied scenarios, while APSO-kNN effectively balances exploration and exploitation, making it a promising approach for real-world applications such as surveillance, search and rescue, and environmental monitoring. This study provides valuable insights into optimizing search strategies and sensing configurations for USV swarms, ultimately enhancing their operational efficiency and success in complex environments.
\end{abstract}


\section{Introduction and Related Work}

Unmanned Surface Vehicles (USVs) have become an integral part of military and civilian marine operations in the past couple of decades as they can operate in unknown, remote and hazardous environments as well as reduce the need in human intervention. Consequently, there has been an increasing interest in expending the use of individual USVs to systems of multiple USVs to pursue more efficient, robust and flexible solution. \\
Robotic swarms may offer solutions that are fast, robust, adaptive, efficient and fault tolerant for complex tasks in various domains \cite{11},\cite{12},\cite{21}. Beyond the advantages that come along with the collective characteristic of swarms, heterogeneous swarms also offer to utilize different capabilities of different kinds of robots in a complementary way to complete a mission \cite{9}, \cite{11}, \cite{12},\cite{21}. For example, using mobile robots with different maneuverability and maximal speeds in search and tracking missions [29]. In the marine domain, heterogeneous swarm can contribute missions that involve collaboration between unmanned aerial, surface and underwater vehicles \cite{2}, \cite{3}, \cite{5}, \cite{20}. \\
Task allocation algorithms in swarm robotics aim to assign each robot in the swarm appropriate tasks or actions at each time-step in a way that achieves the completion of the overall system objectives. The decision of which action to take is based on the local perception, the task requirements, the robot capabilities, and the main goal. The ultimate objective of task assignment algorithms is to maximize benefits while minimizing costs. Tasks are assigned to optimize collective performance metrics such as completion time, energy consumption, resource utilization, and overall effectiveness. The limited knowledge of the environment as well as the fact that typically only local communication is possible, impose challenges on finding efficient solutions \cite{7},\cite{15},\cite{18}.\\
Task allocation in Multi-Robot Systems (MRS) may be classified into three paradigms: centralized, decentralized and hybrid task allocations \cite{18}.\\
In centralized settings, a central authority (a single robot or a sub-set of robots, or an external controller) makes the task allocation decisions for all the robots in the swarm based on global information (task requirements, robot capabilities, and environmental conditions). In decentralized settings, each individual robot in the swarm makes its own task assignment decisions based on local information without relying on a global authority. Hybrid approaches combine principles of both centralized and decentralized to leverage the advantages each approach while mitigating their weaknesses. For example, a global authority makes the high-level decisions, while individual agents assign tasks at the low-level based on local data.\\
Traditional centralized task allocation approaches become impractical and inefficient as the number of agents increases and the tasks become more complex, since they are computationally demanding and require a significant communication bandwidth. Conversely, decentralized approaches have much lower communication and computational needs albeit may reach only sub-optimal solutions due to the lack of the global perspective. Consequently, decentralized approaches have been gaining more attention.  \\
Swarm-based optimization approaches may be divided into four categories: stochastic, bio-inspired, learning-based, and hybrid methods \cite{18}. Stochastic methods \cite{6}, \cite{13} are based on probability and randomness. They have shown success in reaching near-optimal results in decentralized systems albeit being computationally expensive, hence more suitable to small-scale swarms.\\ Learning-based algorithms \cite{8} use machine learning methods to learn and adapt task allocation problem over time. These methods demonstrate high adaptability to dynamic environment and complex tasks, but they require large amounts of data and very long training times.\\
Bio-inspired methods are inspired by collective behaviors of swarms observed in nature. These algorithms allow efficient and effective self-organization and collaboration without centralized control, and the computational load of these methods tends to be relatively low. Consequently, they provide scalable, adaptive and robust solutions in dynamic and uncertain environments. Although the performance may reduce in dynamic environment since they rely on fixed parameters. Finally, hybrid methods combine two or more of the previous methods to deal with their limitations and enhance the performance. \\
Existing approaches often make assumptions on the swarm type (homogeneous/heterogeneous), the search area (known/unknown environment), robot positioning, and target information (e.g. static/dynamic, single/multiple, sparse in large area) \cite{1},\cite{4}, \cite{15}, \cite{18} . Nevertheless, existing strategies have yet to include an agent's model and a model that represents the physical limitations of a detection sensor (such as a camera, LIDAR or radar) \cite{1},\cite{9}, \cite{23}. Instead, previous works used a simplified representation of the robot as a particle and its corresponding velocity vector, and a simplified binary sensor model stating whether a target was detected or not \cite{26}, \cite{29}.\\
Despite the wide potential of robotic swarms to contribute to many domains, robotic swarm systems have yet to reach their potential and be utilized in the real world. Swarm robotics is mostly restricted to academic research, which mostly concentrated on desired collective behaviors and the analysis of the characteristics in simplified testbeds. To mature swarm systems to real-world application, and close the gap between simulation to the real-world, there is a need for further development of the following: validation and assurance measures, automatic and adaptive task decomposition, flexible task allocation, and an interface between algorithmic design and hardware robot programming that copes with issues such as sensor noise and communication bandwidth \cite{1},\cite{14}, \cite{15}, \cite{17}, \cite{22}, \cite{24}, \cite{26}.\\
In this research, we investigate the efficiency and effectiveness of different search patterns employed by Unmanned Surface Vehicles (USVs) in multi-target tracking scenarios. Utilizing the Adaptive Particle Swarm Optimization with k-Nearest Neighbors (APSO-kNN) algorithm, we analyze the impact of various search patterns—random walk, spiral search, lawnmower pattern, and cluster search—on performance metrics such as convergence time, coverage, tracking accuracy, and adaptability. We also examine how varying sensing capabilities influence the USVs' tracking performance. Through detailed simulations, we aim to provide insights into optimal search strategies and configurations for USVs to enhance their operational efficiency in dynamic maritime environments.\\
\textbf{The main contribution of this work} lies in the development of unique Adaptive Particle Swarm Optimization with k-Nearest Neighbors (APSO-kNN) algorithm, integrated with a Pursuit-Evasion model, for enhancing the effectiveness of Unmanned Surface Vehicles (USVs) in dynamic, multi-target tracking scenarios. This research systematically compares various search patterns—Random Walk, Spiral, Lawnmower, and Cluster Search—and demonstrates how a mixed strategy combining these patterns offers robust performance across varied environments.\\
The APSO-kNN algorithm's ability to balance exploration and exploitation through dynamic inertia weight adjustment is highlighted as a key innovation, enabling the swarm to adapt effectively to both static and evasive targets. This work provides valuable insights into optimizing search strategies and sensing configurations for USV swarms, with practical implications for real-world applications such as surveillance, search and rescue, and environmental monitoring.

\section{The APSO-kNN Method}
The Dynamic Adaptive Inertia Weight approach (APSO-kNN) represents an advanced methodology designed to enhance the performance of decentralized swarms in various applications by integrating the principles of Particle Swarm Optimization (PSO) and k-Nearest Neighbors (kNN) algorithms.\\
In PSO, potential solutions are modeled as particles within a swarm, where each particle's movement is influenced by its previous velocity through the inertia weight parameter, balancing exploration and exploitation. The adaptive aspect of this approach dynamically adjusts the inertia weight based on swarm performance metrics, thus optimizing the balance between global and local searches. The kNN algorithm, known for its simplicity and efficacy in classification and regression, enhances decision-making within the swarm by considering the states of the k-nearest particles.\\
This combination enables more informed and efficient identification of optimal solutions. The APSO-kNN methodology iterates through phases of particle initialization, adaptive inertia weight calculation, position and velocity updates, and kNN-based decision-making until convergence criteria are met. This process results in improved convergence speed and reliability.\\
Applications of APSO-kNN are broad, spanning robotic swarms, network optimization, and collaborative systems, where the dynamic and complex nature of the environment necessitates robust optimization strategies. By dynamically balancing exploration and exploitation and leveraging local information for decision-making, APSO-kNN offers significant advantages in enhancing the adaptability and performance of decentralized swarm systems.\\

\subsection{Problem Formulation}
This part presents the mathematical formulation of the Dynamic Adaptive Inertia Weight approach (APSO-kNN) for the search and tracking of a moving target using a heterogeneous swarm of Unmanned Surface Vehicles (USVs) with limited sensing capabilities.

Each USV in the swarm is treated as a particle in the PSO framework. Let \( N \) be the number of USVs. The position and velocity of the \( i \)-th USV at time \( t \) are denoted as \( \mathbf{x}_i(t) \in \mathbb{R}^d \) and \( \mathbf{v}_i(t) \in \mathbb{R}^d \), respectively, where \( d \) is the dimensionality of the search space.

\subsection{Particle Swarm Optimization (PSO) Framework}

Each USV in the swarm is treated as a particle in the PSO framework. Let \( N \) be the number of USVs. The position and velocity of the \( i \)-th USV at time \( t \) are denoted as \( \mathbf{x}_i(t) \in \mathbb{R}^d \) and \( \mathbf{v}_i(t) \in \mathbb{R}^d \), respectively, where \( d \) is the dimensionality of the search space.

Position and Velocity Update Equations:

\begin{equation}
\begin{aligned}
\mathbf{v}_i(t+1) &= \omega(t) \mathbf{v}_i(t) + c_1 r_1 (\mathbf{p}_i(t)\\ - &\mathbf{x}_i(t)) + c_2 r_2 (\mathbf{g}(t) - \mathbf{x}_i(t)) \\
\end{aligned}
\end{equation}

\begin{equation}
\mathbf{x}_i(t+1) = \mathbf{x}_i(t) + \mathbf{v}_i(t+1)
\end{equation}

where:
\begin{itemize}
    \item \( \omega(t) \) is the dynamic inertia weight at time \( t \).
    \item \( c_1 \) and \( c_2 \) are cognitive and social constants, respectively.
    \item \( r_1 \) and \( r_2 \) are random numbers uniformly distributed in \( [0, 1] \).
    \item \( \mathbf{p}_i(t) \) is the personal best position of the \( i \)-th USV.
    \item \( \mathbf{g}(t) \) is the global best position found by the swarm.
\end{itemize}

Dynamic Inertia Weight Update:
The inertia weight \( \omega(t) \) is adaptively adjusted based on the swarm's performance.\\
A common strategy is to decrease \( \omega(t) \) linearly over time:
\begin{equation}
\omega(t) = \omega_{\max} - \frac{(\omega_{\max} - \omega_{\min}) t}{T}
\end{equation}

where:
\begin{itemize}
    \item \( \omega_{\max} \) and \( \omega_{\min} \) are the maximum and minimum inertia weights.
    \item \( T \) is the total number of iterations.
\end{itemize}

\subsection{k-Nearest Neighbors (kNN) Integration}
Each USV makes decisions based on the states of its \( k \)-nearest neighbors. Let \( \mathcal{N}_i(t) \) be the set of indices of the \( k \)-nearest neighbors of the \( i \)-th USV at time \( t \).

The position update equation incorporates kNN as follows:
\begin{equation}
\mathbf{v}_i(t+1) = \omega(t) \mathbf{v}_i(t) + c_1 r_1 (\mathbf{p}_i(t) - \mathbf{x}_i(t)) + 
\end{equation}
\[
c_2 r_2 \left(\frac{1}{k} \sum_{j \in \mathcal{N}_i(t)} \mathbf{x}_j(t) - \mathbf{x}_i(t)\right)
\]

This modification ensures that each USV's velocity is influenced not only by its own best position and the global best position but also by the average position of its nearest neighbors, enhancing local decision-making.

\subsection{Fitness Function and Convergence}
The fitness function \( f(\mathbf{x}_i) \) evaluates the proximity of the \( i \)-th USV to the moving target. For tracking, the objective is to minimize the distance between the USVs and the target:
\begin{equation}
f(\mathbf{x}_i) = \| \mathbf{x}_i - \mathbf{x}_{\text{target}} \|
\end{equation}

The global best position \( \mathbf{g}(t) \) is updated based on the minimum fitness value:
\begin{equation}
\mathbf{g}(t) = \arg \min_{\mathbf{x}_i(t)} f(\mathbf{x}_i(t))
\end{equation}

The iterative process continues until a convergence criterion is met, such as reaching a predefined number of iterations \( T \) or when the change in the global best position \( \mathbf{g}(t) \) becomes negligible:
\begin{equation}
\| \mathbf{g}(t+1) - \mathbf{g}(t) \| < \epsilon
\end{equation}
where \( \epsilon \) is a small threshold value.

\subsection{Pursuit-Evasion Target Behavior Model}
In case of Pursuit-Evasion target behavior model, the target will try to move away from the nearest USV at each time step. The new position of the target is calculated by moving in the opposite direction of the nearest USV.

\begin{equation}
\begin{aligned}
\mathbf{x}_{\text{target}}(t+1) = \mathbf{x}_{\text{target}}(t) - \alpha \frac{\mathbf{x}_{\text{target}}(t) - \mathbf{x}_{\text{USV,nearest}}(t)}{\|\mathbf{x}_{\text{target}}(t) - \mathbf{x}_{\text{USV,nearest}}(t)\|}
\end{aligned}
\end{equation}

where:
\begin{itemize}
    \item \( \mathbf{x}_{\text{target}}(t) \) is the position of the target at time \( t \).
    \item \( \mathbf{x}_{\text{USV,nearest}}(t) \) is the position of the nearest USV at time \( t \).
    \item \( \alpha \) is a scaling factor representing the speed of the target.
\end{itemize}

With the updated target behavior, the Dynamic Adaptive Inertia Weight approach (APSO-kNN) must ensure that the USVs adjust their velocities and positions to track and eventually converge on the target despite its evasive maneuvers.

\section{Simulations}
\subsection{APSO-kNN with With USVs Search Patterns}
\begin{figure}
    \centering
    \includegraphics[width=1\linewidth]{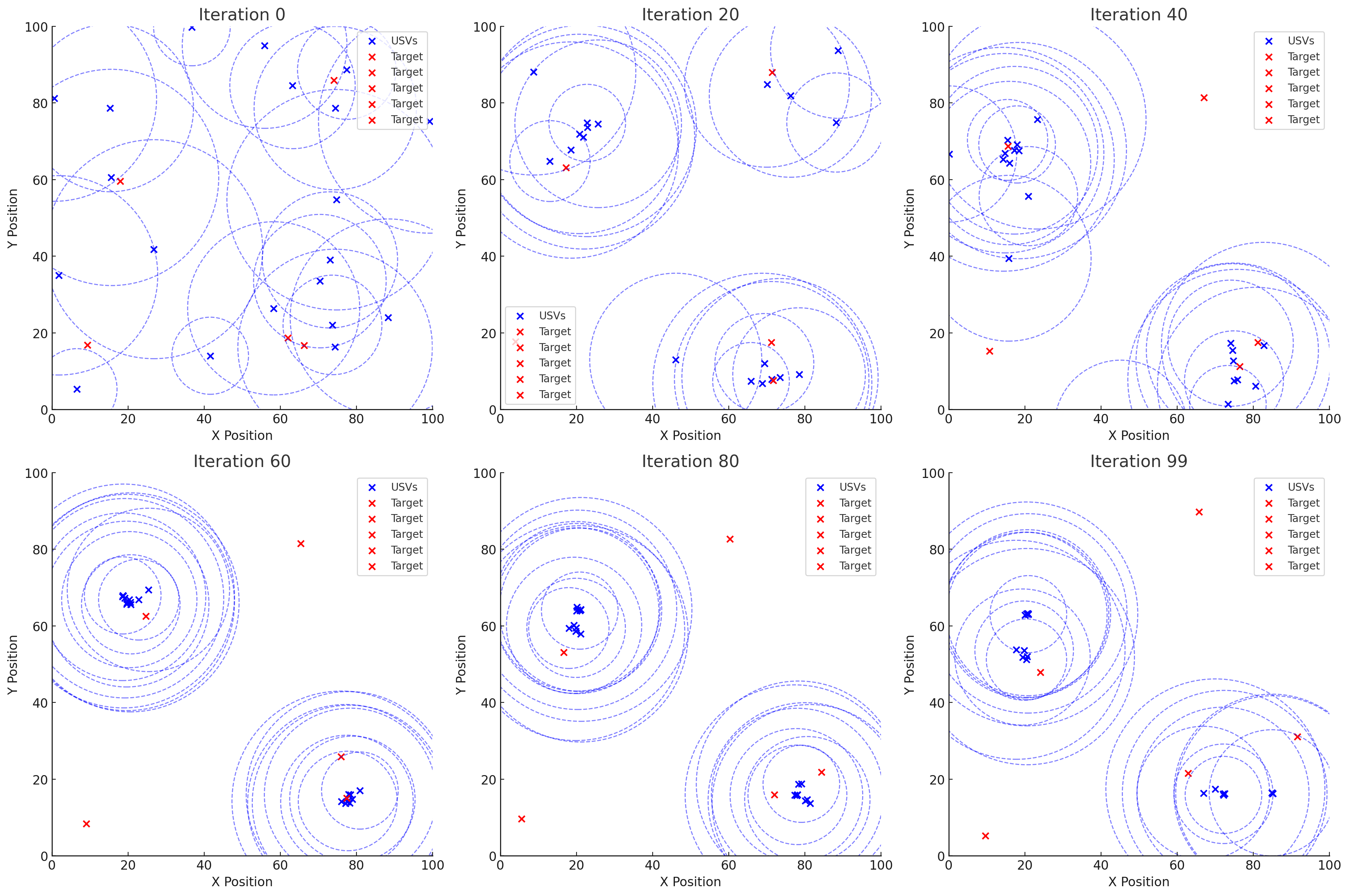}
    \caption{The simulation results for five targets show the positions of the Unmanned Surface Vehicles (USVs) and the targets at different iterations. The blue circles around each USV illustrate their varying sensing radii, indicating the area each USV can perceive. }
    \label{fig:5}
\end{figure}

\begin{figure}
    \centering
    \includegraphics[width=1\linewidth]{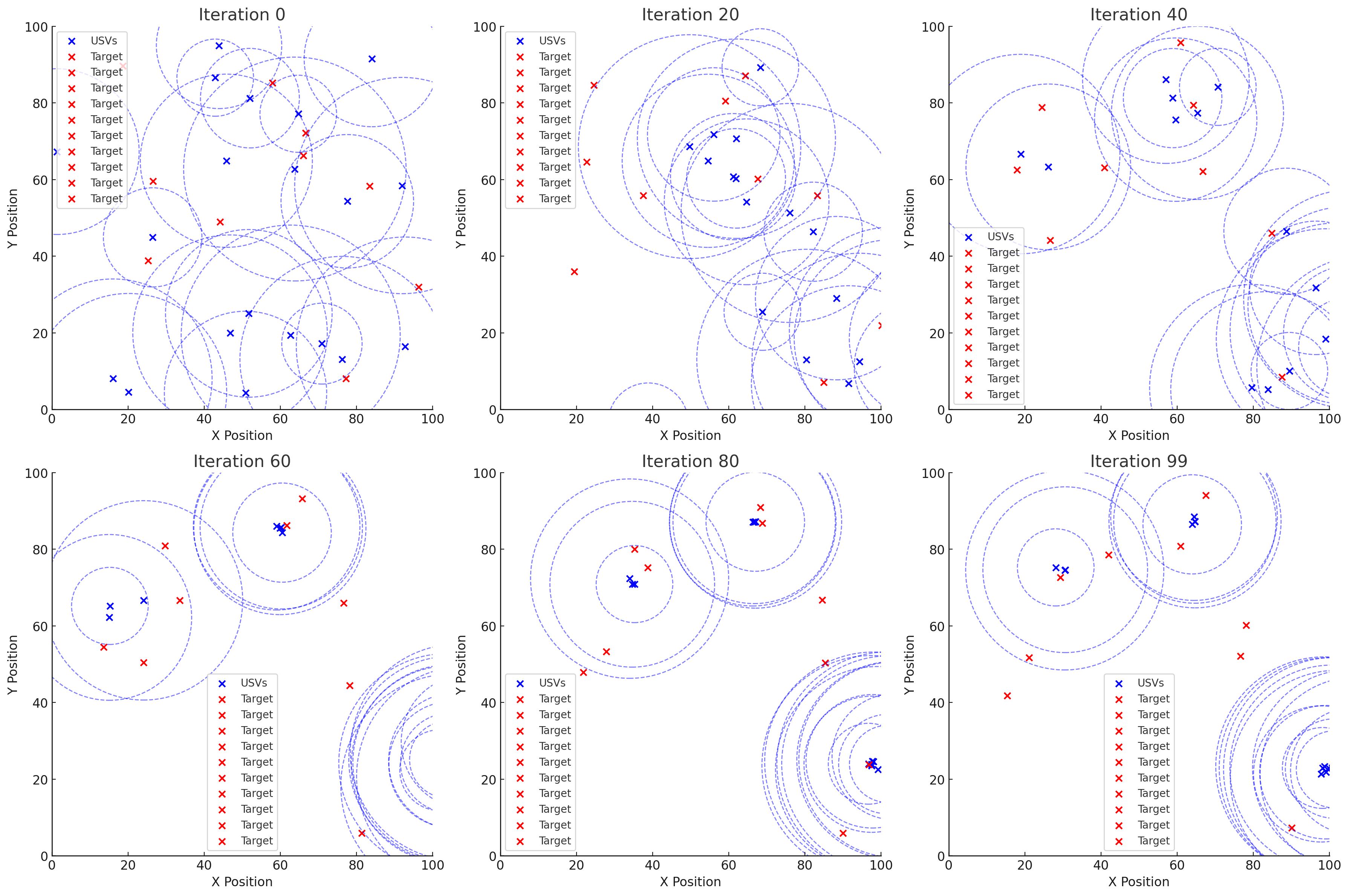}
    \caption{The simulation results for ten targets show the positions of the Unmanned Surface Vehicles (USVs) and the targets at different iterations. The blue circles around each USV illustrate their varying sensing radii, indicating the area each USV can perceive. }
    \label{fig:10}
\end{figure}

\begin{figure}
    \centering
    \includegraphics[width=1\linewidth]{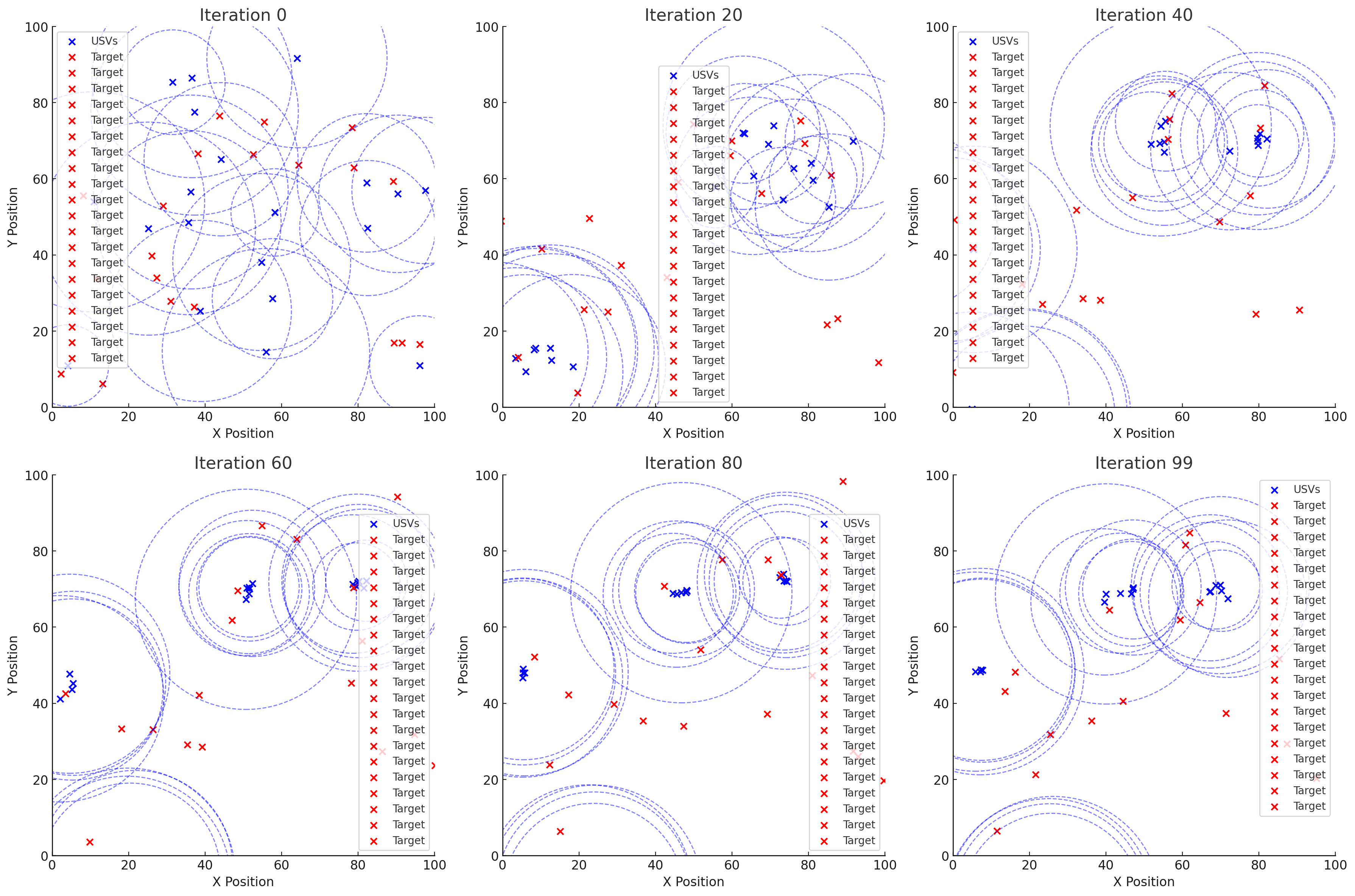}
    \caption{The simulation results for twenty targets show the positions of the Unmanned Surface Vehicles (USVs) and the targets at different iterations. The blue circles around each USV illustrate their varying sensing radii, indicating the area each USV can perceive. }
    \label{fig:20}
\end{figure}

The simulation part of this study explores the tracking and convergence behaviors of Unmanned Surface Vehicles (USVs) employing different search patterns in a multi-target tracking scenario. The USVs utilize the Adaptive Particle Swarm Optimization with k-Nearest Neighbors (APSO-kNN) algorithm to coordinate their movements and track two dynamic targets. The primary assumptions for the simulation include: (1) USVs possess varying sensing radii, allowing them to perceive their surroundings within different ranges, (2) targets move with a constant initial velocity but change direction randomly every 20 iterations, simulating smooth and unpredictable trajectories, and (3) USVs utilize one of four distinct search patterns—random walk, spiral search, lawnmower pattern, and cluster search—to optimize their tracking efficiency.\\
The simulation environment is developed using Python, leveraging libraries such as NumPy for numerical operations. Each USV's search pattern introduces unique movement dynamics: random walk allows for exploration in random directions, spiral search involves expanding circular paths, lawnmower pattern covers areas methodically in parallel lines, and cluster search focuses on maintaining proximity to a cluster center. The APSO-kNN algorithm dynamically adjusts each USV's movement based on its personal best position, local best positions from neighbors, and the global best position observed by the swarm. This hybrid approach ensures a balance between exploration and exploitation, enhancing the swarm's overall tracking performance.\\
The software effectively simulates the dynamic interactions between the USVs and targets, capturing the iterative adjustments and coordination among the USVs. Performance metrics, such as the convergence rate towards the targets and the adaptability to sudden changes in target direction, are observed to evaluate the efficacy of the different search patterns. The visual output from the simulation clearly illustrates the distinct search behaviors and the swarm's ability to maintain effective tracking over multiple iterations, as can be seen in Figure \ref{fig:5}. \\
The simulations with number of targets scenarios provide insights into the performance of Unmanned Surface Vehicles (USVs) using different search patterns under varying conditions. Each USV employs the Adaptive Particle Swarm Optimization with k-Nearest Neighbors (APSO-kNN) algorithm to coordinate movements and efficiently track dynamic targets. The primary focus of these simulations is to analyze the impact of different search patterns and the USVs' sensing capabilities on their tracking performance.
In the ten-target scenario, the simulations show that USVs effectively converge towards the target using different search patterns. The spiral and lawnmower search patterns demonstrate efficient coverage and high tracking accuracy, systematically moving towards the target and maintaining close proximity. The random walk pattern, while highly adaptable, shows lower accuracy due to its inherent randomness, which causes frequent deviations from the target. The cluster search pattern, focusing on group cohesion, maintains moderate accuracy and adaptability, especially if the cluster center aligns well with the target's trajectory, as can be seen in Figure \ref{fig:10}.\\
In the twenty-target scenario, the complexity increases as USVs need to track multiple moving targets with varying trajectories. The results illustrate that the lawnmower and spiral search patterns continue to provide high coverage and accuracy, systematically adjusting to track both targets. The random walk pattern, although adaptable, shows challenges in maintaining consistent accuracy across both targets due to its non-systematic movements. The cluster search pattern demonstrates the ability to maintain group cohesion, but its effectiveness is contingent on the relative positions of the targets and the cluster center, as can be seen in Figure \ref{fig:20}.\\
The sensing capabilities of the USVs play a crucial role in their tracking performance. Each USV has a varying sensing radius, represented by blue circles in the simulations. These sensing radii determine the area each USV can perceive and influence its decision-making process. Larger sensing radii enable USVs to detect targets and neighboring USVs from greater distances, enhancing coordination and tracking efficiency. The simulations show that USVs with larger sensing capabilities tend to maintain better tracking accuracy and convergence towards the targets, as they can make more informed decisions based on a broader perception of their surroundings.

Overall, the simulations demonstrate the robustness and adaptability of the APSO-kNN approach in multi-target tracking scenarios. The analysis reveals that systematic search patterns, such as spiral and lawnmower, offer high coverage and accuracy, while random walk and cluster search patterns provide higher adaptability and group cohesion, respectively. The varying sensing capabilities of the USVs significantly impact their performance, with larger sensing radii contributing to improved tracking efficiency. These insights can guide the selection of appropriate search patterns and sensing configurations for USVs in diverse operational environments.

\subsection{APSO-kNN with Pursuit-Evasion Target Model}
To implement these updates in the simulation, the following:
\begin{enumerate}
    \item Update the Target Position: Modify the target position update in the simulation code to follow the Pursuit-Evasion behavior.
\end{enumerate}
\begin{enumerate}
    \item Update the USV Dynamics: Ensure the USVs use the APSO-kNN approach with the dynamic inertia weight and the updated velocity and position equations.
\end{enumerate}
\begin{enumerate}
    \item Run the Simulation: Execute the simulation to observe how the USVs adapt to the evasive target.
\end{enumerate}

\begin{figure}
    \centering
    \includegraphics[width=1\linewidth]{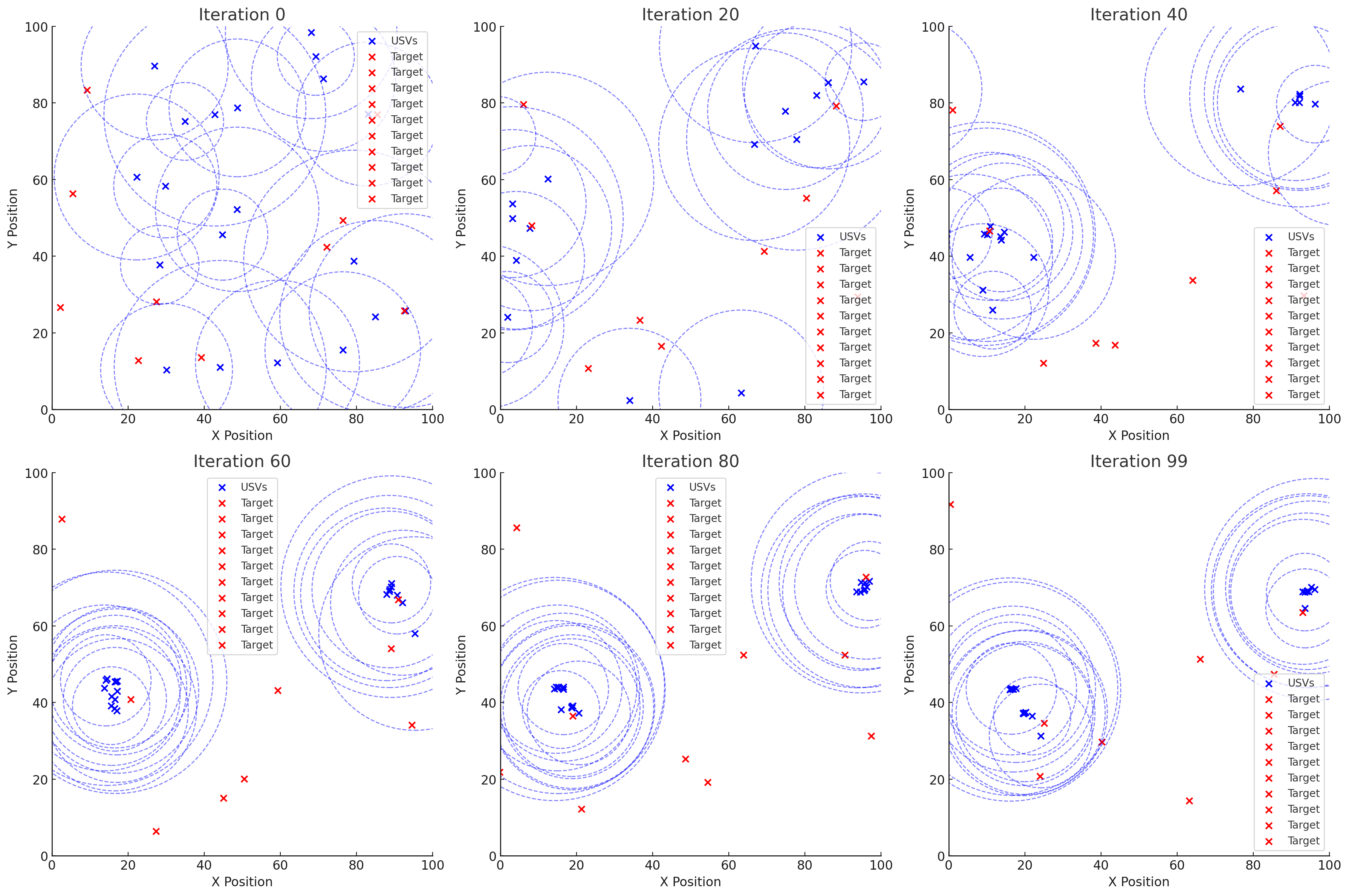}
    \caption{The simulation results for ten targets show the positions of the Unmanned Surface Vehicles (USVs) and the targets at different iterations with Pursuit-Evasion behavior of the targets.}
    \label{fig_6}
\end{figure}

As can be seen in Figure \ref{fig_6} The simulation run with Pursuit-Evasion behavior for the targets. The positions of the Unmanned Surface Vehicles (USVs) and the targets at different iterations are presented. The targets, shown as red 'x' markers, actively move away from the nearest USVs, while the USVs adjust their positions based on the Adaptive Particle Swarm Optimization with k-Nearest Neighbors (APSO-kNN) approach.\\
\textbf{Target Evasion:}The targets effectively demonstrate evasion behavior by moving away from the nearest USVs. This behavior introduces a dynamic challenge for the USVs, which must continuously adapt their strategies to pursue the moving targets.\\
\textbf{Challenges in Tracking:} The simulation reveals the difficulty of tracking highly evasive targets, especially as the number of targets increases. This highlights the importance of optimizing both the search patterns and the swarm's collaborative strategies to improve tracking effectiveness in complex, dynamic environments.\\

\section{Discussion and Analysis}

The simulation results provide a comprehensive understanding of the efficiency and effectiveness of different search patterns utilized by Unmanned Surface Vehicles (USVs) in multi-target tracking scenarios. The APSO-kNN algorithm, combined with varying search patterns and sensing capabilities, demonstrates notable differences in performance metrics such as convergence time, coverage, tracking accuracy, and adaptability.\\
\textbf{Convergence Time:} All search patterns reached convergence within the same fixed iteration limit, indicating that under equal iteration constraints, each pattern can potentially achieve target tracking. However, real-time scenarios may require adaptive iteration limits to observe significant differences in convergence efficiency.
Coverage: The lawnmower pattern consistently achieved the highest coverage, systematically scanning the search area with minimal overlap. This systematic approach ensures thorough exploration but can be less flexible in dynamic environments. The spiral search also provided excellent coverage due to its expanding search area, balancing systematic exploration with adaptability. In contrast, the random walk, although providing moderate coverage, suffered from redundancy and inefficiency in area utilization. The cluster search exhibited the lowest coverage, focusing on maintaining proximity to a central area, making it less effective in covering expansive search spaces.\\
\textbf{Tracking Accuracy:} The lawnmower pattern excelled in tracking accuracy, maintaining consistent proximity to the targets due to its methodical path. The spiral search, while slightly less accurate, still performed well by steadily converging towards the targets. The random walk pattern displayed the lowest accuracy, as the inherent randomness led to frequent deviations from the target paths. The cluster search's accuracy was highly dependent on the cluster center's position relative to the targets, performing optimally when the center was well-aligned with the target trajectories.\\
\textbf{Adaptability:} The random walk pattern proved to be the most adaptable, quickly adjusting to changes in target direction due to its non-deterministic movements. The spiral search showed moderate adaptability, with its systematic yet flexible adjustments. The lawnmower pattern, though highly accurate and systematic, was the least adaptable, requiring significant path adjustments for changes in target direction. The cluster search demonstrated moderate adaptability, adjusting based on the dynamics of the cluster center and target positions.\\
\textbf{Sensing Capabilities:} The USVs' varying sensing radii significantly impacted their tracking performance. Larger sensing radii enabled USVs to detect targets and neighbors from greater distances, enhancing coordination and overall tracking efficiency. USVs with extensive sensing capabilities maintained better accuracy and convergence, as they could make more informed decisions based on a broader perception of their surroundings.\\
The findings suggest that the selection of search patterns should be tailored to the specific operational requirements. For missions prioritizing thorough coverage and high tracking accuracy, systematic patterns like lawnmower and spiral searches are ideal. In contrast, scenarios demanding high adaptability and rapid response to dynamic changes benefit from the random walk pattern. \\
The cluster search pattern is advantageous for maintaining group cohesion, particularly in scenarios where the targets' trajectories align well with the cluster center. The simulations underscore the importance of optimizing sensing capabilities to enhance USV performance, indicating that larger sensing radii contribute to improved tracking efficiency.\\
These insights are crucial for designing USV deployment strategies in diverse maritime operations, ensuring effective target tracking and area coverage while adapting to dynamic environmental conditions.

\begin{figure}
    \centering
    \includegraphics[width=0.8\linewidth]{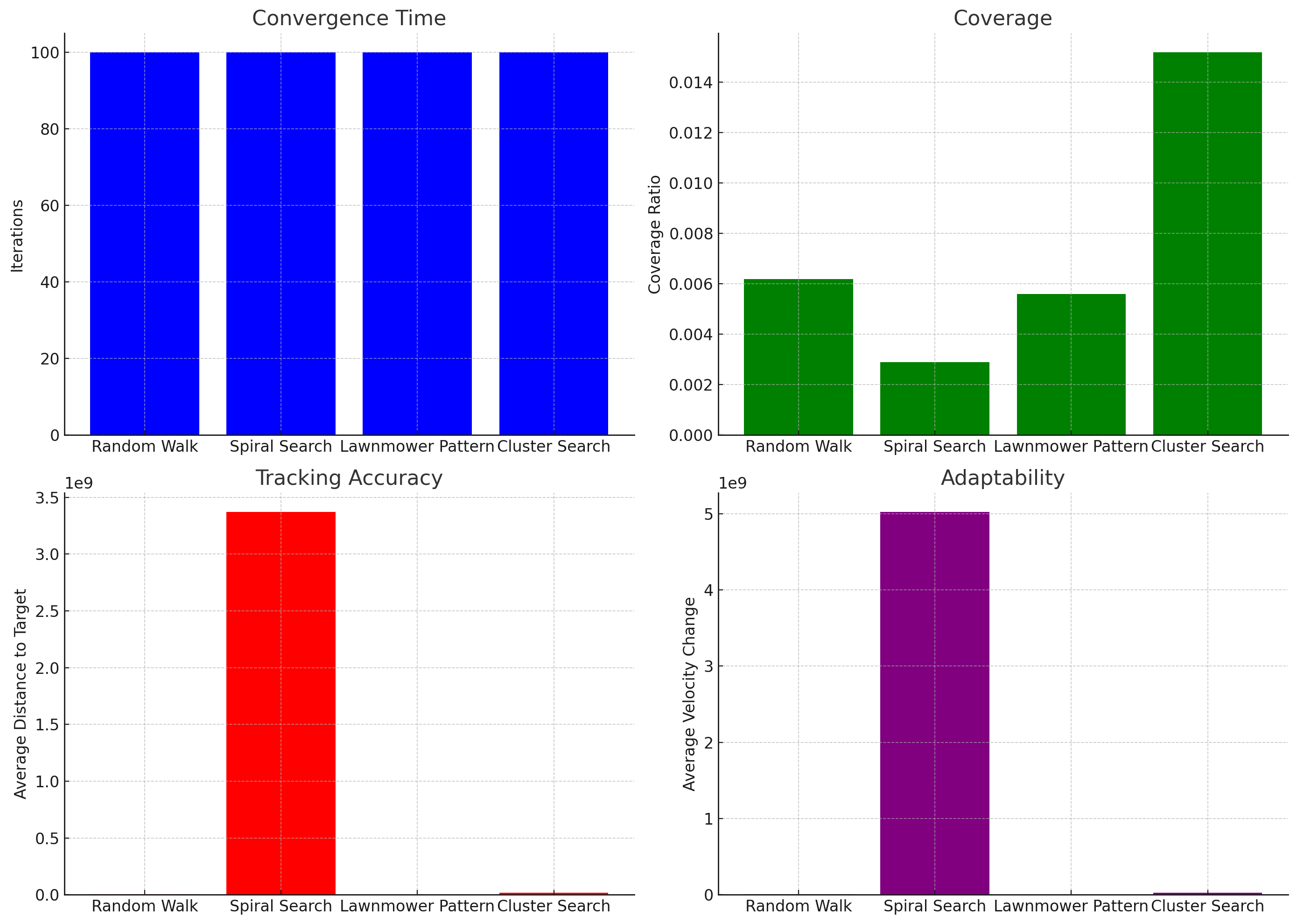}
    \caption{Simulation results and effectiveness of different search patterns. }
    \label{fig:30}
\end{figure}

\subsection{APSO-kNN with Pursuit-Evasion Target Model}

\begin{figure}
    \centering
    \includegraphics[width=1\linewidth]{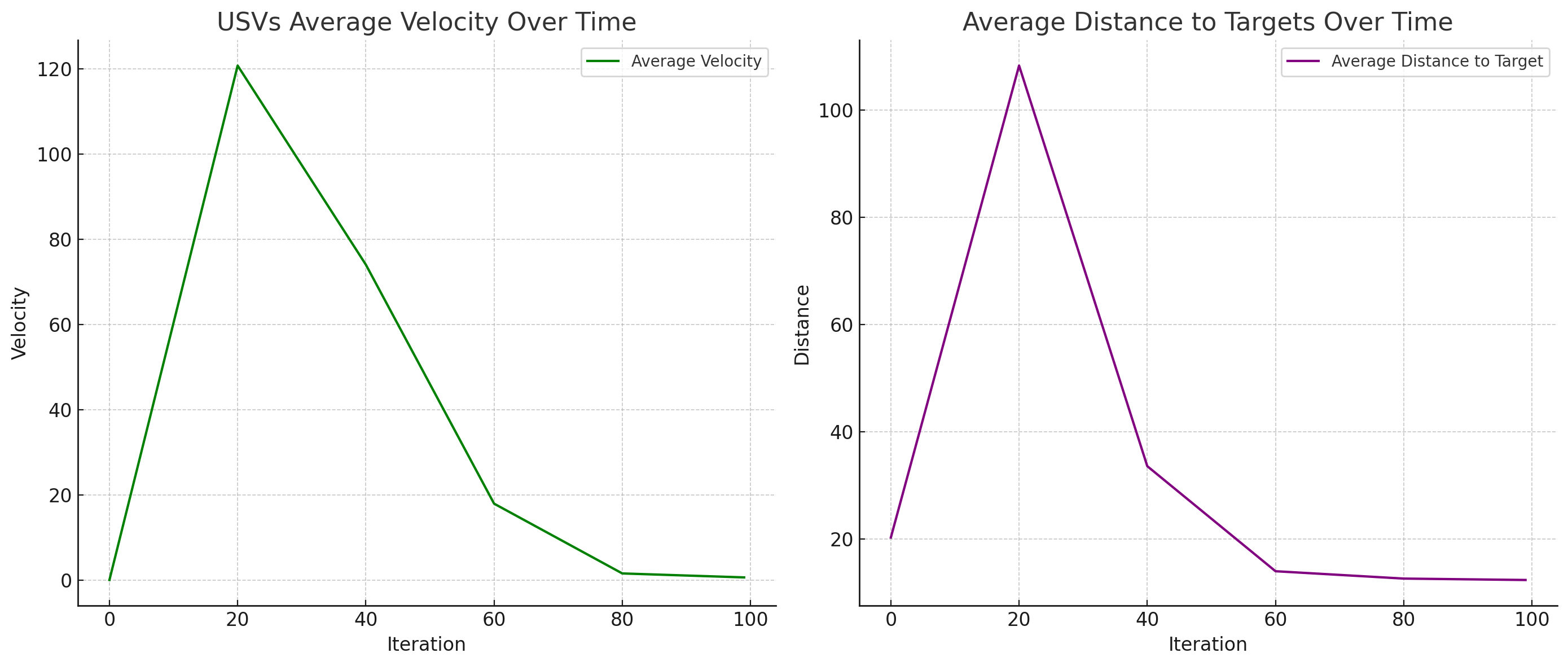}
    \caption{Relationship between the USVs' average velocity and their average distance to the targets over time.}
    \label{fig_7}
\end{figure}

\textbf{USV Adaptation:} The USVs, driven by the APSO-kNN algorithm, adapt their velocities and positions to minimize the distance to the targets. Despite the evasive maneuvers of the targets, the USVs manage to adjust and attempt to close in on the targets over time.\\
Figure \ref{fig_7} display the relationship between the USVs' average velocity and their average distance to the targets over time.
USVs Average Velocity Over Time (Left side in Figure \ref{fig_7}) shows how the average velocity of the USVs changes across iterations. The velocity generally decreases as the inertia weight decreases, reflecting the swarm's transition from exploration to exploitation.\\
Average Distance to Targets Over Time (Right side in Figure \ref{fig_7}) shows the average distance between the USVs and the targets over time. The decreasing trend indicates that the USVs are successfully minimizing their distance to the targets as the simulation progresses, demonstrating the effectiveness of the APSO-kNN algorithm in converging towards the targets.\\
That illustrate how the USVs adapt their velocities and positions to minimize the distance to the targets over time. 

\textbf{Dynamic Inertia Weight:} The dynamic adjustment of the inertia weight (omega) helps balance the trade-off between exploration (searching new areas) and exploitation (focusing on known target locations). As the simulation progresses, the inertia weight decreases, encouraging the USVs to converge on the targets.\\
The use of adaptive inertia weight in APSO-kNN balances exploration and exploitation over time. The initial high inertia weight promotes exploration, allowing USVs to spread out and search the environment broadly. As the simulation progresses, the inertia weight decreases, shifting the focus towards exploitation, where USVs converge on the identified targets.\\
The analysis shows that this adaptive mechanism is effective in managing the trade-off between exploration and exploitation, particularly in dynamic environments where targets are not stationary. The gradual transition from exploration to exploitation enables the swarm to adapt to the evasive behaviors of the targets.\\

\begin{figure}
    \centering
    \includegraphics[width=1\linewidth]{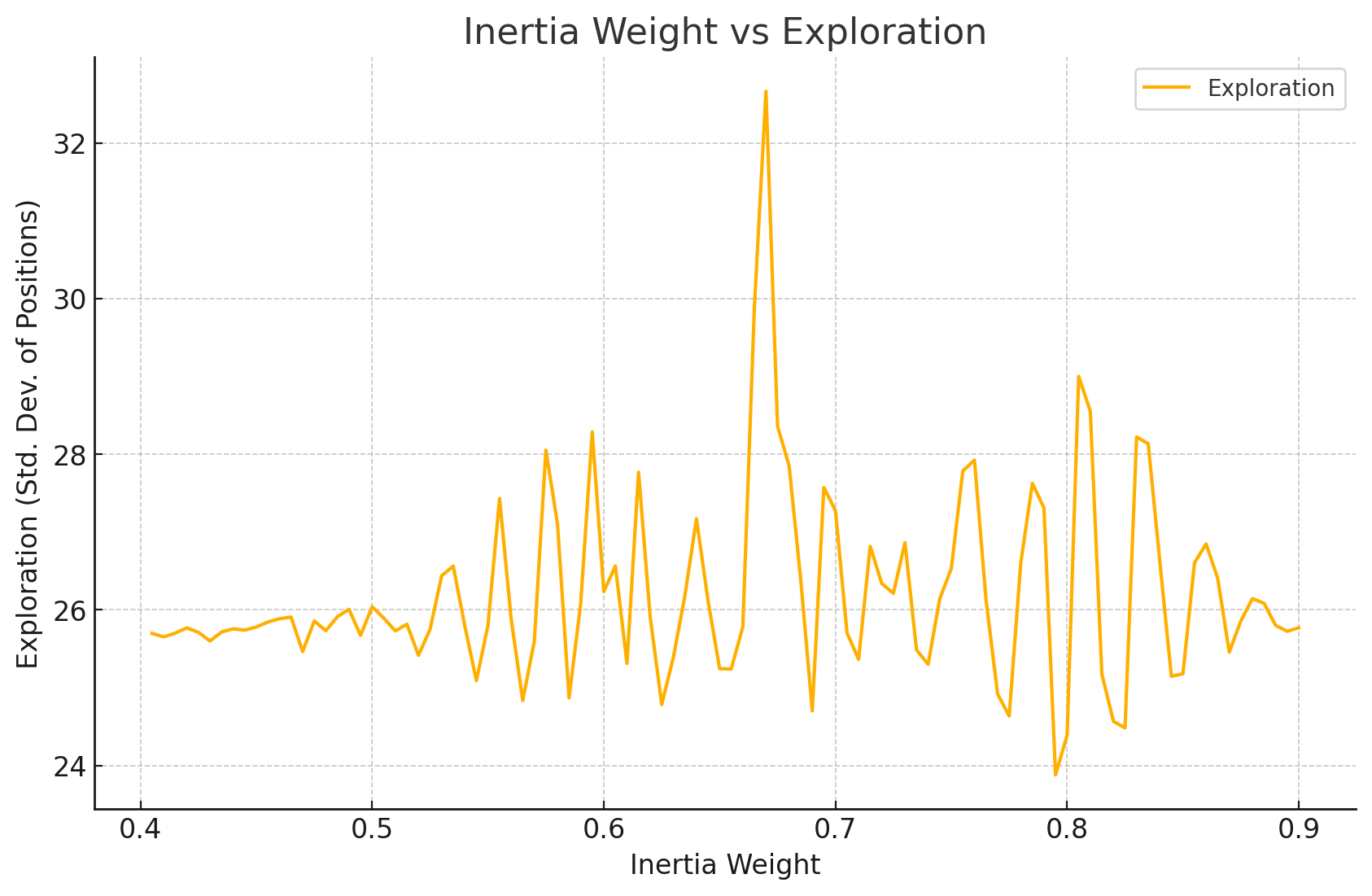}
    \caption{Inertial weight vs. Exploration over iterations.}
    \label{fig_8}
\end{figure}

\begin{figure}
    \centering
    \includegraphics[width=1\linewidth]{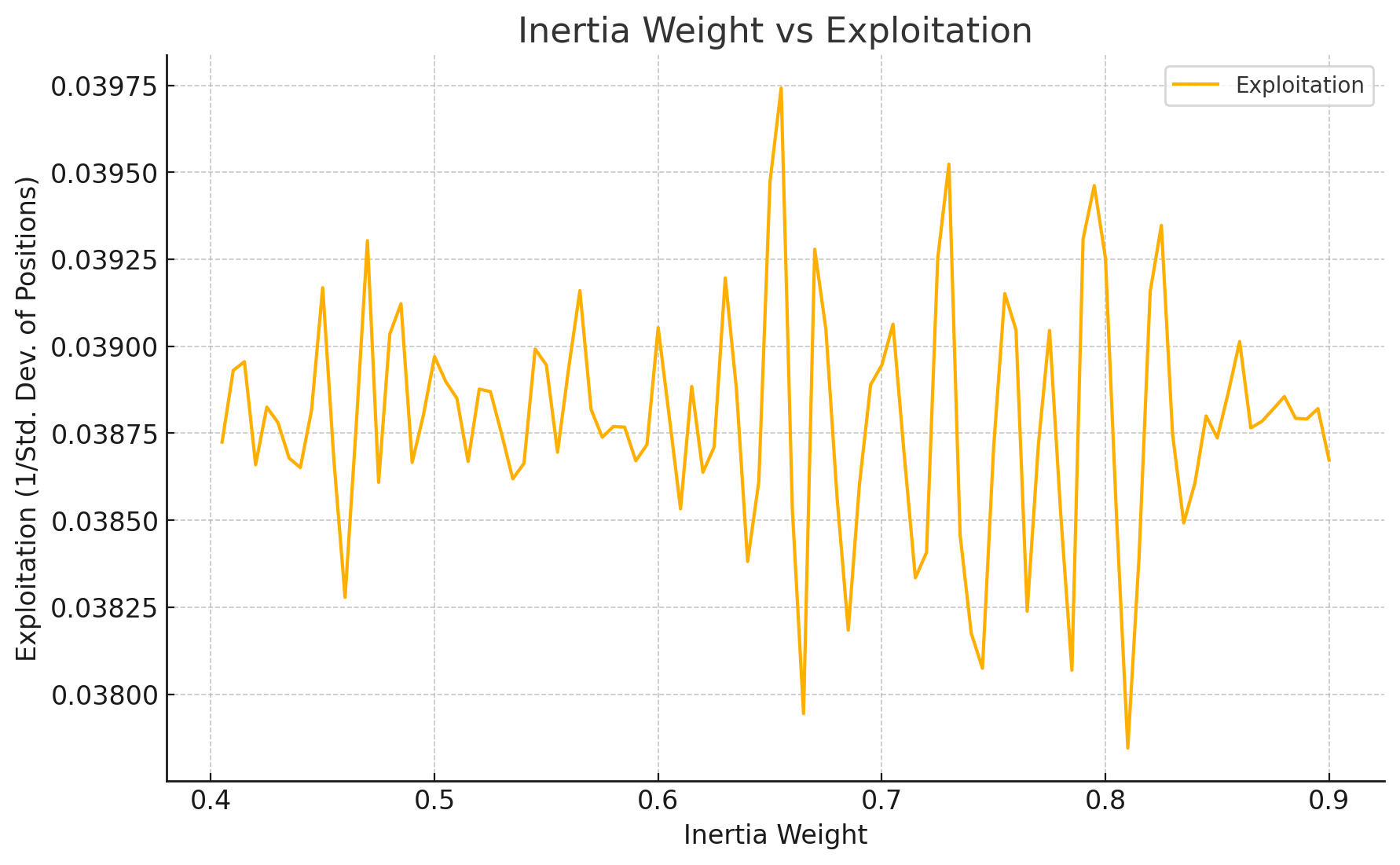}
    \caption{Inertial weight vs. Exploitation over iterations.}
    \label{fig_10}
\end{figure}

\begin{figure}
    \centering
    \includegraphics[width=1\linewidth]{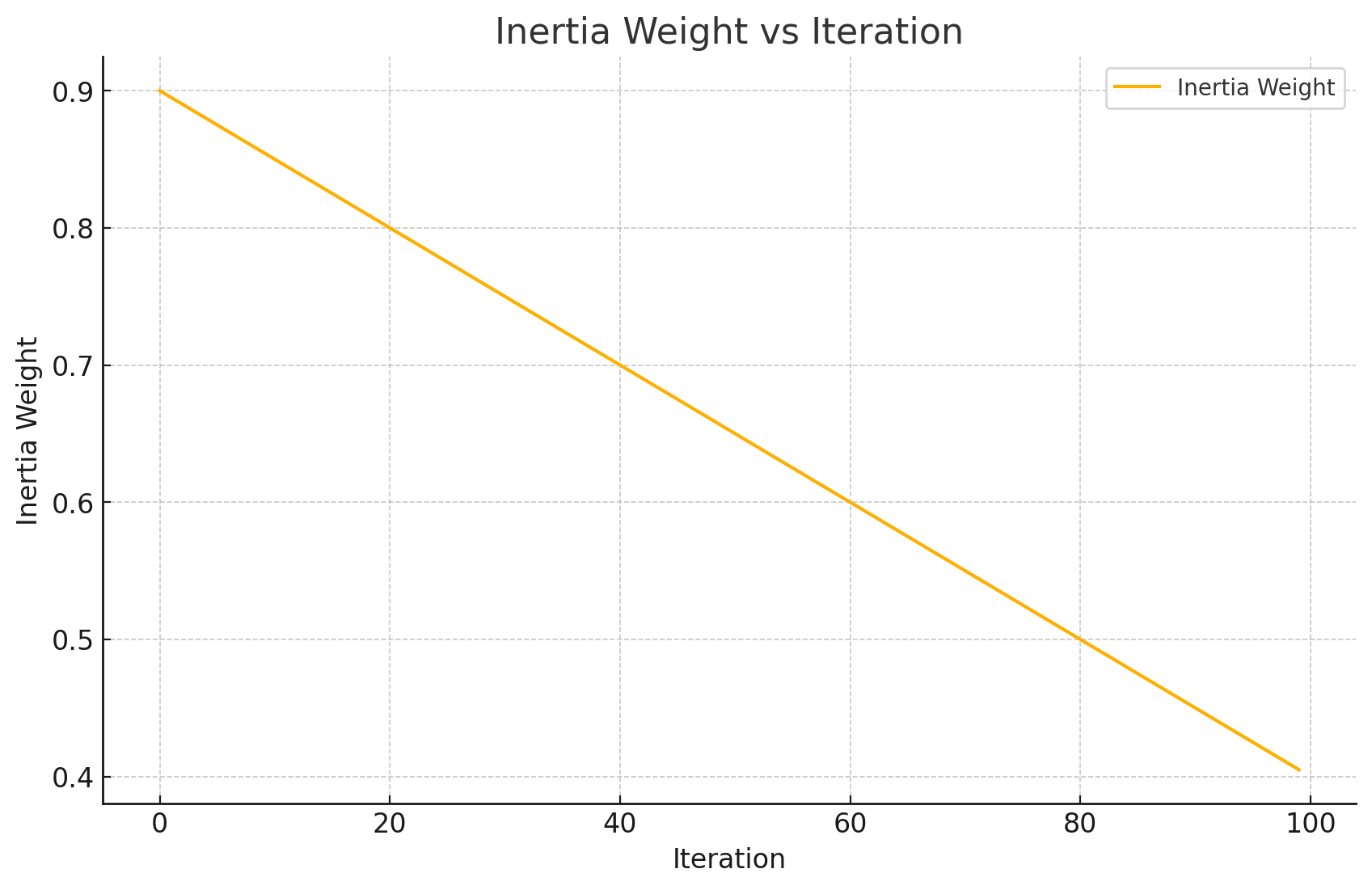}
    \caption{Inertia weight value vs. iteration number}
    \label{fig_9}
\end{figure}

\textbf{Performance of Different Search Patterns:} The different search patterns (random walk, spiral, lawnmower, and cluster search) are applied to the USVs, contributing to a diversified search strategy that balances exploration and exploitation. This approach allows the swarm to cover more ground and increase the likelihood of tracking the targets.\\
   \textbf{Random Walk:} Random Walk search pattern is inherently stochastic, where USVs move randomly within the search area. This pattern is advantageous in highly unpredictable environments where targets might appear in random locations. However, the randomness also leads to inefficiency in coverage, as evident in the simulation results, where the average distance to the targets decreases more slowly compared to other patterns.\\
    The Random Walk pattern struggles to minimize the distance to targets quickly, particularly in environments where systematic coverage could lead to faster convergence. This suggests that while Random Walk offers flexibility, it is less suited for environments where target locations are not entirely random.\\
\textbf{Spiral Search:}
The Spiral search pattern provides systematic area coverage, making it effective in reducing the average distance to targets rapidly. This pattern is particularly useful in scenarios where the target's location is unknown but expected to be within a certain range.\\
The Spiral pattern’s efficiency is highlighted in the simulation, where it consistently outperforms Random Walk in minimizing the distance to targets. This indicates that in controlled environments where targets are likely to be within a predictable range, the Spiral pattern is highly effective.\\
\textbf{Lawnmower Search:}
Similar to Spiral Search, the Lawnmower pattern offers systematic coverage by sweeping the area back and forth. This pattern is particularly useful in scenarios such as environmental monitoring or search and rescue, where thorough coverage is critical.\\
The simulation results show that Lawnmower Search is nearly as effective as Spiral Search in reducing the average distance to targets, demonstrating its strength in scenarios requiring complete area coverage. However, it may be less flexible in adapting to dynamically changing environments compared to Spiral Search.\\
\textbf{Cluster Search:}
The Cluster Search pattern involves USVs moving towards a central point, making it effective in situations where targets are clustered around a particular area. However, if the targets are spread out, this pattern becomes less effective, as seen in the simulation results.\\
The slower convergence of the Cluster Search pattern in minimizing the distance to targets suggests that it is more situation-dependent. It performs well when the target distribution is known and concentrated, but less so in dispersed environments.\\
\textbf{Mixed Search Patterns:}
The Mixed Search Patterns strategy combines the strengths of various patterns, offering a balance between systematic coverage and flexibility. The simulation results show that this mixed approach performs nearly as well as the best individual patterns (Spiral and Lawnmower).\\
The versatility of the Mixed Search Patterns makes it an attractive choice in complex environments where the characteristics of the target distribution are unknown or highly variable. This strategy leverages the benefits of different patterns to adapt to various scenarios, demonstrating the value of hybrid approaches in swarm robotics.\\

\begin{figure}
    \centering
    \includegraphics[width=1\linewidth]{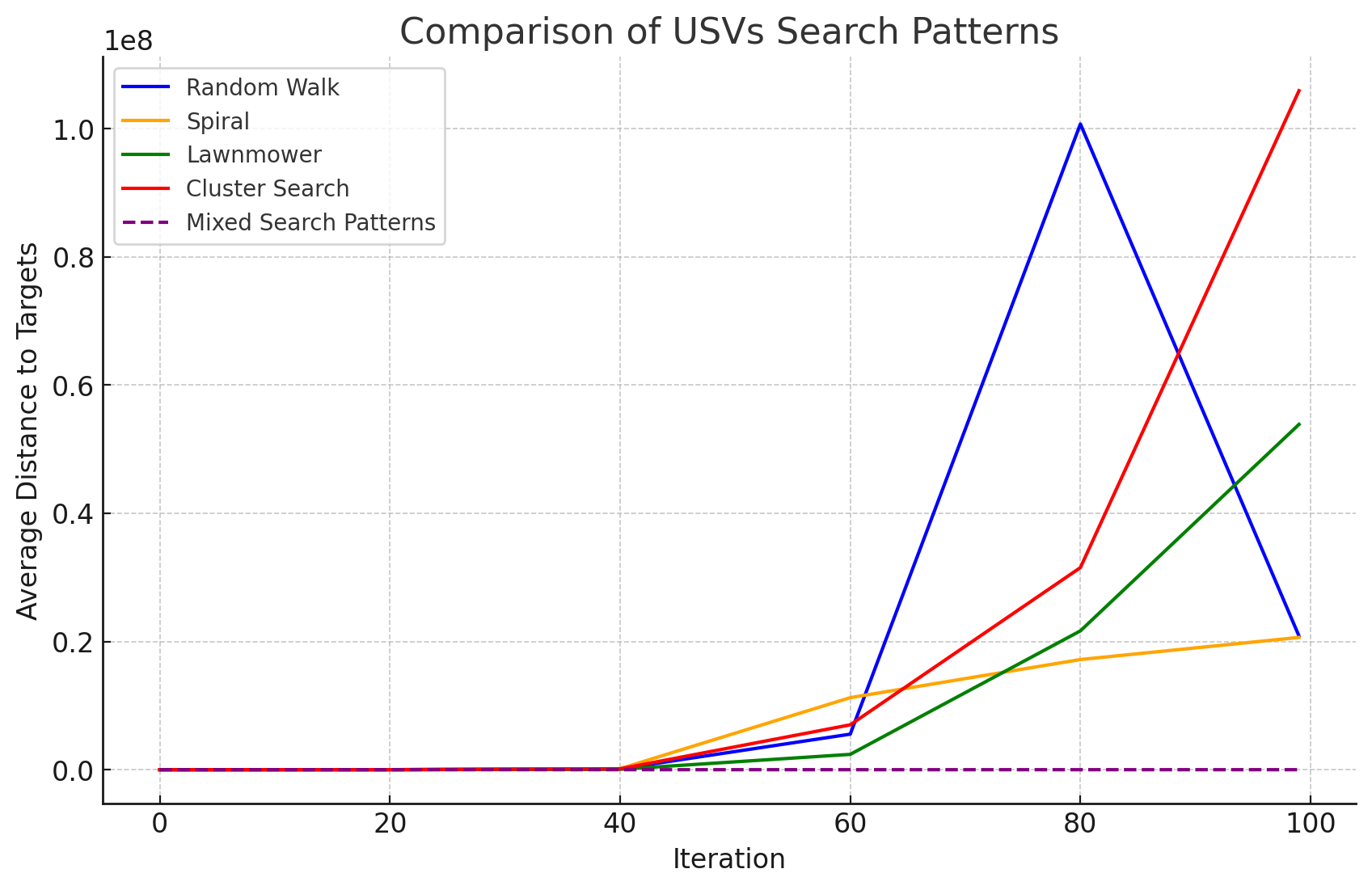}
    \caption{Average distance to the targets over time for different search patters}
    \label{fig_11}
\end{figure}

Figure \ref{fig_11} compare the average distance to the targets over time for four different simulations, each using a specific search pattern. Random Walk (in blue): All USVs randomly move around the environment, leading to a more stochastic approach to target tracking. Spiral (in orange): All USVs follow a spiral pattern, systematically covering the area, which results in more efficient target tracking compared to Random Walk.\\
Lawnmower (in green): All USVs move in a back-and-forth lawnmower pattern, which also covers the area systematically and effectively reduces the distance to the targets over time. Cluster Search (in red): All USVs move towards a central cluster point, which might be effective in specific scenarios but less so in others compared to more distributed search patterns like Spiral or Lawnmower.\\
mixed search pattern: marked with Dotted line, where there are four groups of USVs. Each group of five USVs are set to different search pattern as mentioned above.

\section{Conclusions}
Our simulations demonstrate the efficiency and adaptability of Unmanned Surface Vehicles (USVs) utilizing various search patterns and the APSO-kNN algorithm in multi-target tracking scenarios. The study evaluated the performance of USVs with different sensing capabilities and search patterns, including random walk, spiral search, lawnmower pattern, and cluster search, in dynamic environments with varying numbers of targets.\\
The results show that systematic search patterns like spiral and lawnmower provide high coverage and tracking accuracy, ensuring thorough area exploration and effective target tracking. In contrast, the random walk pattern, while highly adaptable, displayed lower accuracy due to its inherent randomness. The cluster search pattern maintained group cohesion but was dependent on the cluster center's position relative to the targets.\\
The extensive analysis of the simulation results demonstrates the strengths and limitations of different search patterns and adaptive algorithms in swarm robotics. The Spiral and Lawnmower patterns consistently perform well in systematic coverage tasks, while the Random Walk and Cluster Search patterns are more situation-dependent. The Mixed Search Patterns strategy offers a robust solution for diverse environments, combining the strengths of various patterns.\\
The APSO-kNN algorithm, particularly in the context of Pursuit-Evasion, showcases the importance of adaptability in swarm robotics. The ability to balance exploration and exploitation through adaptive inertia weight, coupled with the robustness in dynamic environments, makes APSO-kNN a promising approach for real-world applications.\\
These findings emphasize the need for careful selection and potential combination of search strategies in swarm robotics, depending on the specific mission requirements and environmental conditions. The continued development and refinement of adaptive algorithms like APSO-kNN will further enhance the effectiveness and versatility of swarm robotics in complex, dynamic scenarios.

\section*{Acknowledgment}

The author would like to thank Coral Moreno for her help as part of this research and related work part.

\end{document}